\documentclass[letterpaper, 10 pt, conference]{ieeeconf}  

\IEEEoverridecommandlockouts                              

\usepackage{graphicx} 
\usepackage{times} 
\usepackage{amsmath} 
\usepackage{amssymb}  
\usepackage{booktabs}
\usepackage{multirow}
\usepackage[hidelinks]{hyperref}
\usepackage{algorithm}
\usepackage{algpseudocode}
\usepackage{subcaption}
\usepackage{xcolor}
\usepackage{cite}

\newcommand{\bbm}{\begin{bmatrix}}
	\newcommand{\ebm}{\end{bmatrix}}
\def \ssm#1{\left[ #1 \right]_{\!\times}}

\makeatletter
\newcommand*\titleheader[1]{\gdef\@titleheader{#1}}
\AtBeginDocument{%
	\let\st@red@title\@title
	\def\@title{%
		\bgroup\normalfont\large\centering\@titleheader\par\egroup
		\vskip1.5em\st@red@title}
}
\makeatother
\titleheader{© 20XX IEEE.  Personal use of this material is permitted.  Permission from IEEE must be obtained for all other uses, in any current or future media, including reprinting/republishing this material for advertising or promotional purposes, creating new collective works, for resale or redistribution to servers or lists, or reuse of any copyrighted component of this work in other works.}

\title{\LARGE \bf
	DeRO: Dead Reckoning Based on Radar Odometry With Accelerometers Aided for Robot Localization
}

\author{Hoang Viet Do, Yong Hun Kim, Joo Han Lee, Min Ho Lee, and Jin Woo Song$^*$
	\thanks{This work was supported by Unmanned Vehicles Core Technology Research and Development Program through the National Research Foundation of Korea (NRF), Unmanned Vehicle Advanced Research Center (UVARC) funded by the Ministry of Science and ICT, the Republic of Korea (No. 2020M3C1C1A01086408 and NRF-2023M3C1C1A01098408). ($^*$Corresponding author: Jin Woo Song.)}%
	\thanks{All authors are with the Intelligent Navigation and Control Systems Laboratory (iNCSL), School of Intelligent Mechatronics Engineering, and the Department of Convergence Engineering for Intelligent Drone, Sejong University, Seoul 05006, Republic Of Korea. {\tt\small \{hoangvietdo}; {\tt\small yhkim}; {\tt\small dlwngks12}; {\tt\small mhleee\}}{\tt\small @sju.ac.kr}; {\tt\small jwsong@sejong.ac.kr}}%
}

\begin{document}
	
	\maketitle
	\thispagestyle{empty}
	\pagestyle{empty}

	\begin{abstract}
		In this paper, we propose a radar odometry structure that directly utilizes radar velocity measurements for dead reckoning while maintaining its ability to update estimations within the Kalman filter framework.
		Specifically, we employ the Doppler velocity obtained by a 4D Frequency Modulated Continuous Wave (FMCW) radar in conjunction with gyroscope data to calculate poses.
		This approach helps mitigate high drift resulting from accelerometer biases and double integration.
		Instead, tilt angles measured by gravitational force are utilized alongside relative distance measurements from radar scan matching for the filter's measurement update.
		Additionally, to further enhance the system's accuracy, we estimate and compensate for the radar velocity scale factor. 
		The performance of the proposed method is verified through five real-world open-source datasets.
		The results demonstrate that our approach reduces position error by 62\% and rotation error by 66\% on average compared to the state-of-the-art radar-inertial fusion method in terms of absolute trajectory error.
	\end{abstract}

	
	\section{INTRODUCTION}
	In recent decades, achieving precise and reliable localization has emerged as a paramount challenge for advanced autonomous robots.
	A widely employed approach involves integrating an Inertial Measurement Unit (IMU) with supplementary sensors to address this challenge.
	The rationale behind this fusion lies in the fact that while IMUs provide high-rate but short-term accuracy, additional drift-free sources are required.
	Among these methods, the fusion of IMU data with that from a Global Navigation Satellite System (GNSS) stands out as the most widely used, with rich developmental history \cite{groves}.
	However, GNSS signals are not always available, as in tunnels or indoor environments, and their reliability degrades in densely built urban areas due to operational constraints \cite{gnss-denied}.
	
	To date, the common GNSS-denied navigation strategies have primarily relied on visual sensors such as cameras or Light Detection and Ranging (LiDAR) sensors. 
	While the fusion of these sensors with an IMU yields significant advancements \cite{vins-review, liosam2020shan}, they do not offer a comprehensive solution for all environmental conditions.
	Specifically, cameras are susceptible to failure in low-light conditions, such as darkness or low-textured scenes.
	One potential approach to addressing this challenge involves the utilization of thermal or infrared cameras \cite{thermal}; still, these alternatives are ineffective in the absence of sufficient temperature gradients.
	Conversely, LiDAR operates independently of light conditions but experiences degradation when operating in the presence of fog or smoke \cite{lidar-fail}.
	\begin{figure}[t]
		\centering
		\includegraphics[scale=0.38]{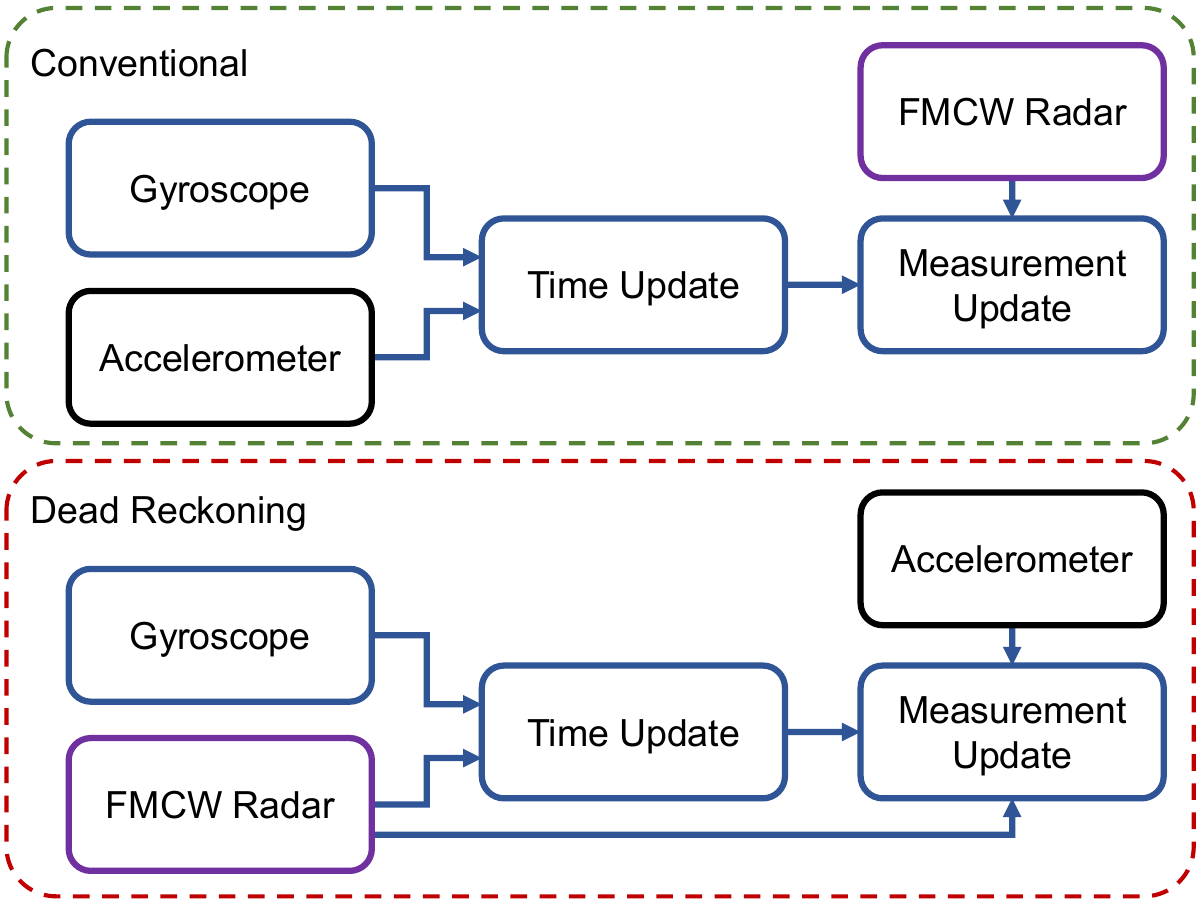}
		\caption{Block diagram illustrates the distinction between conventional RIO and DR-based RO structures.}
		\label{diff_structure}
	\end{figure}

	Instead of relying solely on visual sensors like cameras or LiDAR, another viable option for GNSS-denied navigation is the utilization of a 4-dimensional (4D) Frequency-Modulated Continuous Wave (FMCW) radar \cite{radar-survey-1}.
	Radar offers distinct advantages, particularly in adverse weather conditions such as fog or smoke, where the previous mentioned sensors may fail.
	Moreover, modern FMCW radars are characterized by their affordability, lightweight and compact size, making them suitable for small-size robotic application with space-limitation such as drones.
	Another significant advantage of 4D FMCW radar is its ability to provide Doppler velocity in addition to range, azimuth, and elevation angle information of targets.
	This feature holds the potential for achieving superior localization results compared to other sensor systems.
	
	Considerable efforts have recently been dedicated to integrating an IMU with a 4D FMCW radar for indoor localization \cite{radar-survey-2}.
	Given that radar offers 4D point cloud measurements of targets, two primary approaches can be employed to determine the robot's poses.
	The first approach, known as the instantaneous method, utilizes Doppler velocities and angles of arrival from a single scan to estimate ego velocity \cite{instant}.
	The second approach involves scan matching between two radar scans, enabling not only localization but also mapping \cite{4d-radar-slam}.
	However, due to challenges such as signal reflection, ghost targets, and multi-path interference, this method remains demanding and requires extensive tuning and computational resources.
	On the other hand, in applications where mapping is unnecessary, the former approach proves to be more robust and reliable.
	
	To the best of the author's knowledge, existing literature predominantly treats radar as an auxiliary sensor within a loosely coupled framework.
	This methodology is commonly referred to as radar inertial odometry (RIO).
	In this framework, radar measurements typically serve the purpose of updating the state estimation, often implemented through an extended Kalman filter (EKF) \cite{DoerMFI2020}.
	%
	%
	Meanwhile, an inertial navigation system (INS) algorithm integrates acceleration to derive velocity, and then integrates again to determine the platform's position \cite[Ch. 11]{titterton2004strapdown}.
	However, this conventional methodology is notably sensitive to the accuracy of accelerometer bias estimation.
	Even a modest error in bias estimation, as small as 10\%, can result in significant position errors \cite{dr}.

	To address these challenges, the dead reckoning (DR) fusion technique has been proposed and implemented \cite{dr}.
	When a velocity sensor such as an odometer is available, it replaces the accelerometer for position estimation.
	This substitution offers several advantages.
	Firstly, it completely prevents significant errors caused by accelerometers biases (typically from a low-grade IMU), while also eliminating the need for a second integration step to calculate position estimation.
	Additionally, when relying on accelerometer measurements, compensating for gravitational forces becomes necessary \cite{titterton2004strapdown}.
	Inaccuracies in determining this constant can lead to notable errors due to the effect of double integration.
	Moreover, conventional INS may include additional attitude errors caused by vibrations, such as those encountered in drones or legged robots.
	Vibration induces acceleration biases in IMU measurements, whereas it is typically considered as noise in velocity sensors.
	%
	Generally, handling noise is considered more straightforward than addressing biases.
	%
	
	Motivated by the aforementioned advantages and existing empirical support for the DR technique \cite{dr}, this paper introduces a DR-based structure for radar odometry (RO), as depicted in Fig. \ref{diff_structure}.
	In this framework, we integrate the 4D FMCW radar and gyroscope as the core navigation algorithm, while treating the accelerometers as aided sensors.
	We argue that radar is highly suitable for the DR structure, as it completely eliminate the side-slip error associated with traditional odometer-based DR approaches in automotive applications.
	Furthermore, the versatility of the DR technique can be extended to non-wheeled robots (e.g., drones, legged robots and ships) when radar is employed.
	
	To sum up, our primary contributions of this article lie in the following aspects:
	\begin{enumerate}
		\item A framework of \textbf{De}ad reckoning based on \textbf{R}adar \textbf{O}dometry (DeRO) with accelerometers aided is presented.
		In this pipeline, the radar's ego velocity and angular velocity measurement from gyroscope are used for the filter's time update.
		Simultaneously, the radar's range measurement and tilt angles computed by accelerometers contribute to the estimation correction under stochastic cloning concept.
		\item We estimate and compensate for the 3D radar velocity scale factor to optimize the performance of the proposed system.
		\item We conduct a comparative analysis with state-of-the-art RIO approaches using the same open-source dataset to validate the effectiveness of our proposed system.
		\item We implement the proposed method using C++ in robot operating system 2 (ROS 2) \cite{ros2} and make our source code openly available\footnote{\url{https://github.com/hoangvietdo/dero}.} for the benefit of the research community.
	\end{enumerate}

	The rest of this article is organized as follows.
	In Section \ref{sec:related-work}, we review related work in the context of RIO and DR.
	Definitions on coordinate frames and mathematical notations along with DR-based radar odometry mechanization are given in Section \ref{sec:dr}.
	We develop the stochastic cloning based indirect EKF starting from the state-space definition in Section \ref{sec:ekf}.
	Next, Section \ref{sec:results} describes the real-world experiment results to demonstrate our proposed framework.
	Finally, Section \ref{sec:conclusion} concludes this article.
	
	\section{RELATED WORKS} \label{sec:related-work}
	\subsection{Dead Reckoning}
	The dead reckoning method has been extensively researched and popularly adopted as a straightforward and computationally effecient solution for odometer/INS/GNSS integrated system in GNSS-denied environments \cite{dr-1, dr-2, dr-3}.
	Simulation and field test results have consistently shown the effectiveness of this approach over the conventional INS-based system \cite{dr,YU2018936}.
	To further improve its robustness, adaptive rules for the EKF were introduced by \cite{dr-adaptive} to handle measurement outliers and dynamic model disturbances.
	In addition, Takeyama \textit{et al.} \cite{dr-camera} extended the traditional DR-based structure by incorporating a monocular camera to improve heading estimation accuracy.
	
	Concurrently, many researchers have focused on improving odometer quality by estimating and compensating for the odometer's scale factor \cite{jhj}.
	Park \textit{et al.} \cite{dr} addressed nonholomic constraints of the wheeled vehicle to suspend the odometer side-slip errors.
	The authors also utilized tilt angles calculated from gravitational force measurements to further enhance roll and pitch angles estimation.
	The same approach was also explored in \cite{dr-3}.
	
	One thing we notice from these previous researches is that, most of the DR-based application has primarily been applied to land vehicles, with limited extensions to other platforms such as drones.
	This restriction stems from the availability of odometers in wheeled vehicles, their cost-effectiveness, and they require minimal effort to compute the vehicle's body velocity.
	Moreover, it is not straightforward to directly measure body velocity in other non-wheeled applications.
	In contrast to previous approaches, our work employs a 4D FMCW radar instead of an odometer, thus overcoming this limitation.
	
	\subsection{Radar Inertial Odometry}
	According to \cite{radar-survey-2}, one of the earliest contributions to radar inertial odometry is the EKF RIO \cite{DoerMFI2020} introduced by Doer and Trommer.
	The key achievement was the expansion of previous research on 2D radar odometry with motion constraints into the realm of 3D.
	Specifically, they integrated inertial data from IMUs with 3D radar-based ego velocity estimates using EKF framework.
	Additionally, 3-point Random sample consensus (RANSAC) estimator has been applied to deal with radar noisy point clouds.
	This impressive results formed the foundation for their subsequent studies in this field.
	In \cite{doer-no-yaw}, they employed stochastic cloning technique to estimate the IMU-radar extrinsic calibration matrix, further refining the accuracy of the EKF RIO.
	Next, they introduced a novel method for calculating the relative heading angle from radar ranges measurement under Manhattan world assumptions \cite{doer-yaw}.
	Expanding on their work, in \cite{doer-thermal}, Doer and Trommer extended the RIO pipeline by incorporating both monocular and thermal cameras to tackle extremely challenging visual conditions.
	Recently, they have further enhanced the system by adding a GNSS receiver \cite{DoerAeroConf2022}.
	Ng \textit{et al.} \cite{c-rio} proposed a continuous-time framework for fusing an IMU with multiple heterogeneous and asynchronous radars.
	
	In contrast to the Doppler-based approaches mentioned above, where only a single radar scan is utilized, the technique of calculating relative pose between two scans via scan matching is less favored due to the sparsity and noise nature of radar measurements.
	Michalczyk \textit{et al.} \cite{scan-matching} presented a hybrid method that update the estimation by leveraging both ego velocity measurement and matched 3D points obtained from scan matching using stochastic cloning technique \cite{stochastic}.
	In \cite{radar-msckf}, they introduced a novel \emph{persistent landmarks} concept, which serves to increase the accuracy of scan matching.
	Almalioglu \textit{et al.} \cite{ndt} used Normal Distributions Transform (NDT) to determine the optimal transformation between corresponding points in consecutive radar scans.
	Subsequently, this measurement was integrated into the EKF RIO without incorporating Doppler velocity measurements.

	Despite the utilization of radar measurements, radar remained classified as an assisted sensor in the aforementioned studies.
	Our approach, on the other hand, considers radar as both the primary sensor and an aided sensor concurrently.
	Particularly, Doppler-based measurement will primarily be used for position calculation, while range measurements will serve to update the estimation.
	A key motivation behind this configuration is the superior accuracy of velocity estimates derived from radar compared to those from an IMU, especially in high dynamic motions.
	As a result, in the conventional framework, such as RIO, the velocity estimate provided by the INS typically converges towards the velocity estimate derived from radar.
	
	\section{DEAD RECKONING USING RADAR ODOMETRY} \label{sec:dr}
	\subsection{Coordinate Frame and Mathematical Notations}
	Throughout the forthcoming, we define the navigation (global) frame $\{n\}$ as the local tangent frame fixed at the starting point of the body frame $\{b\}$ of a robot. In this frame, the axes are oriented towards the north, east, and downward directions.
	The IMU frame is coincident with $\{b\}$, wherein its axes pointing forward, right and down, respectively.
	The FMCW radar frame, denoted as $\{r\}$, is oriented towards the forward, left, and upward directions, with its origin positioned at the center of the transmitter antenna.
	We use superscript and subscript to indicate a reference frame and a target frame, respectively.
	When referring to a specific time instance, a time index is appended to the subscript.
	For example, the notation $x_{b,k}^n$ signifies a vector $x$ of $\{b\}$ at a time step $k$ expressed in $\{n\}$.
	Moreover, we adopt the symbol $C_b^n$ to express the direction cosine matrix that transform any vector from frame $\{b\}$ to frame $\{n\}$.
	Note that $C_b^n$ belongs to the special orthogonal group in three dimensions, denoted as $SO(3)$.

	For mathematical representation, we employ  $\mathbb{R}^n$ and $\mathbb{R}^{m\times n}$ to denote the set of $n$-dimensional Euclidean space and the set of $m \times n$ real matrices, respectively.
	Scalars and vectors are written as lower case letters $x$, matrices as capital letters $X$, and their transpose as $X^{\top}$.
	The notation $I_n$ indicates an identity matrix whose dimension is $n\times n$.
	An $m \times n$ zeros matrix is denoted by $0_{m\times n}$.
	A general multivariate Gaussian distribution of a random vector $x$ with mean $\mu$ and covariance matrix $P$ is denoted as $x \thicksim {\mathcal{N}}\big(\mu, P \big)$.

	\subsection{System Overview}
	\begin{figure}[t]
		\centering
		\includegraphics[scale=0.335]{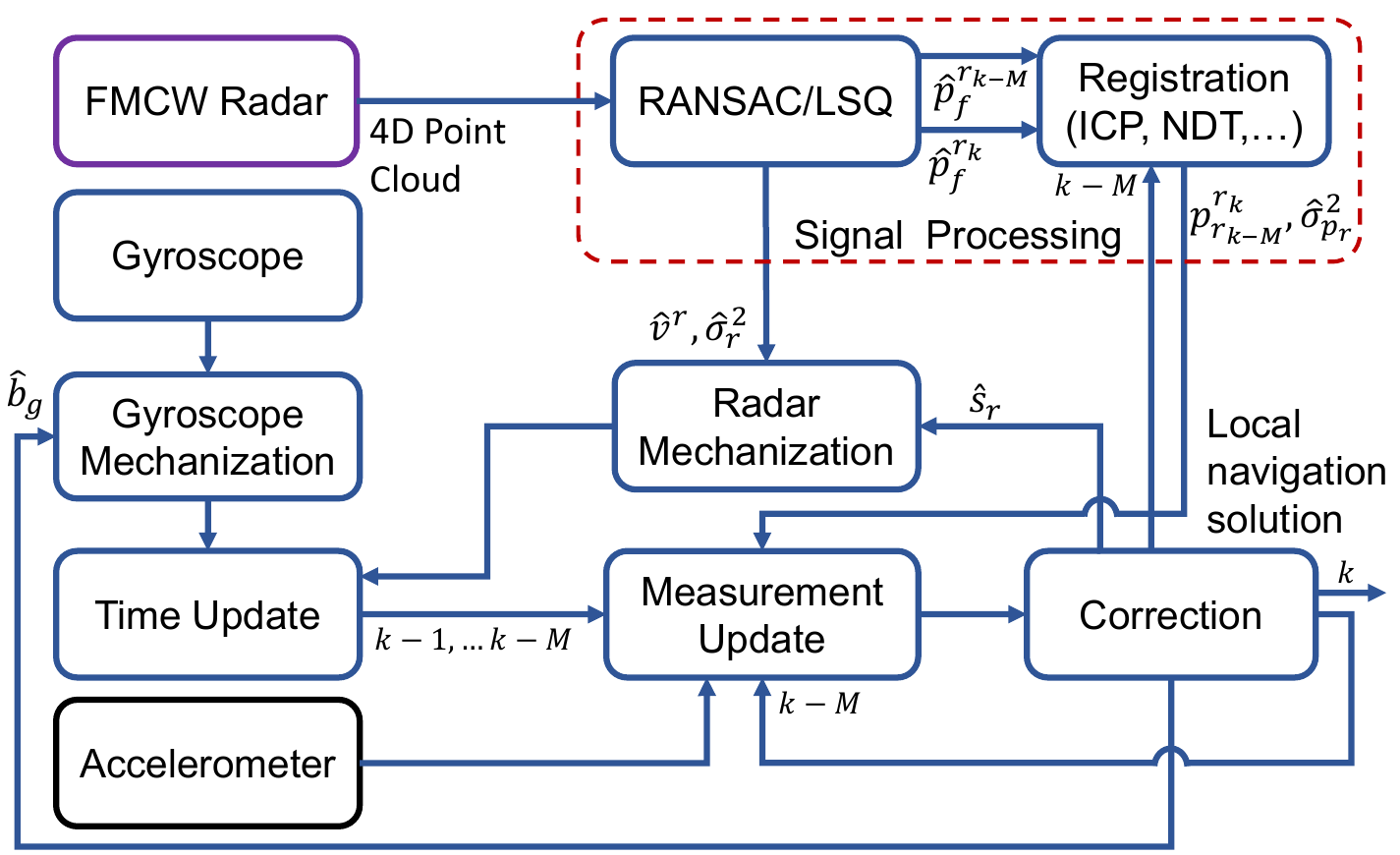}
		\caption{Overview block diagram of the proposed method.}
		\label{overview}
	\end{figure}

	The overview of the proposed DeRO is depicted in Fig. \ref{overview}.
	In this study, we apply a stochastic cloning based error-state (indirect) EKF (SC-IEKF) to fuse data from an IMU and a 4D FMCW radar.
	The raw point clouds obtained from radar are first fed into an outlier rejection algorithm.
	Here, we employ a simple RANSAC-based least square (RANSAC/LSQ) method introduced in \cite{DoerMFI2020}.
	The outputs of this algorithm include the estimated ego velocity $\hat{v}^r$ along with its corresponding error covariance $\hat{Q}$, as well as a set of inlier point clouds.
	These velocity estimates and covariance matrices are used for radar-based DR mechanization in conjunction with gyroscope data to compute poses and the filter's time update.
	Subsequently, the inlier point cloud set is merged with the inlier set from the one at $M$ preceding steps to perform scan matching to calculate the relative distance estimate $\hat{p}^{r,k}_{r,{k-M}}$ with error covariance $\hat{R}$ between these two radar scans.
	This information will update the estimation.
	
	It is important to note that we operate under the assumption of a static environment.
	%
	%
	Chen \textit{et al.} \cite{drio} have demonstrated that the accuracy of ego velocity estimation from radar degrades rapidly in the presence of moving objects within the scene.
	However, in the case that the number of static object dominates the moving object, then RANSAC might be sufficient to remove the influence of moving objects.
	Thus, our assumption can be relative relaxed.
	Addressing scenarios with a high number of moving objects falls outside the scope of this article.
	
	\subsection{Radar-based Dead Reckoning Mechanization} \label{subsec:radar_mechanization}
	Given the target's radial velocity $v_d^r$, one could estimate the radar velocity $v^r$ using RANSAC/LSQ (for more details, refer to \cite{DoerMFI2020}).
	Based on Euler's equations for rigid body motion, we establish the following relationship \cite{DoerMFI2020}
	\begin{align} \label{euler}
		v^r &= C_b^{r} \bigg( C_n^{b} v_b^n + \ssm{\omega^b} p_{r}^b \bigg),
	\end{align}
	where $v_b^n$ is the body's velocity expressed in the navigation frame, $\omega^b$ is the angular velocity of the body frame, and $p_r^b$ is the position of radar with respect to body frame.
	Here, we assume that the calibration parameters $C_b^r$ and $p_r^b$ are known and obtained through a well-calibrated process.

	Since radar velocity measurement is used to calculate the localization information, we transform \eqref{euler} as follows
	\begin{align} \label{radar_mechanization}
		v_b^n &=  C_b^n C_r^b v^r - C_b^n \ssm{\omega^b} p_{r}^b.
	\end{align}
	We can now integrate this velocity to obtain the position of the body frame expressed in $\{n\}$ frame, denoted as $p_b^n$.
	Note that the rotation matrix $C_b^n$ can be computed from the gyroscope's angular velocity through integration \cite{titterton2004strapdown}.
	
	\section{STOCHASTIC CLONING INDIRECT EXTENDED KALMAN FILTER} \label{sec:ekf}
	The fundamental concept of an SC-based filter is to consider the correlation between the current state estimate and the previous state estimate.
	This consideration is necessary when previous estimates are used to calculate essential information for filtering purposes, such as estimating the measurement residual in an EKF.
	This correlation can be effectively accounted for by augmenting the previous and current estimates into a new single state, allowing the new state error covariance to encapsulate their correlation.
	\subsection{State Augmentation}
	Inspired by the odometer-based DR \cite{dr}, we define the DR state vector as
	\begin{align} \label{INS-state}
		x_{dr} = \bbm p_{b}^{n^\top} & q_{b}^{n^\top} & b_g^\top & s_r^\top \ebm^\top \in \mathbb{R}^{13},
	\end{align}
	where the unit quaternion (4-tuple) $q_b^n \in \mathbb{R}^4$ with the norm constraint $\Vert q_b^n \Vert = 1$ is referenced to the global frame.
	The vector $b_g$ represents the gyroscope biases expressed in the body frame, while $s_r \in \mathbb{R}^3$ denotes the radar scale factor.
	
	As we use error-state formulation of the EKF, the error-state variable is used instead of the full-state variable $x$.
	Given the estimate $\hat{x} = \textrm{E} \big[ x \big]$, with the exception of the quaternion, we define the error-state as follows
	\begin{align} \label{error_def}
		\delta x = \hat{x} - x.
	\end{align}
	For the attitude part, the error quaternion $\delta q$ is given as
	\begin{align} \label{quaternion_error_def}
		q_b^n = \delta q_b^n \otimes \hat{q}_b^n,
	\end{align}
	where $\delta q_b^n = [ 1 \,\,\, 0.5 \delta \Psi_b^{n^\top} ]^\top$ under small angle assumption and $\delta \Psi_b^n$ is a small misalignment angle.
	The mathematical symbol $\otimes$ denotes the quaternion multiplication.
	Using this definition, one could get
	\begin{align} \label{cbn_error_def}
		\delta C_b^n = C \big( \delta q_b^n \big) \approx I_3 + \ssm{\delta \Psi_b^n}.
	\end{align}
	Therefore, the DR's error-state vector is given by
	\begin{align}
		\delta x_{dr} = \bbm \delta p_{b}^{n^\top} & \delta \Psi_{b}^{n^\top} & \delta b_g^\top & \delta s_r^\top \ebm^\top \in \mathbb{R}^{12}.
	\end{align}
	
	To deal with the relative measurement from radar, we apply the renowned stochastic cloning technique \cite{stochastic}.
	In particular, given a $M$ step window, we augment the INS state vector with a cloned state from $k-M$ previous steps as follows
	\begin{align}
		\delta x_k = \bbm \delta x_{dr,k}^\top &  \delta p_{b,k-M}^{n^\top} & \delta \Psi_{b,k-M}^{n^\top} \ebm^\top \in \mathbb{R}^{18}.
	\end{align}
	
	\subsection{System Model}
	We now define the mathematical model of the gyroscope and radar sensors as
	\begin{subequations}
		\begin{align}
			\bar{\omega}^b &= \omega^b + b_g + n_g \label{gyroscop_measurement} \\
			\bar{v}^r &= \textrm{diag}( s_r )^{-1} v^r + n_r, \label{radar_measurement}
		\end{align}
	\end{subequations}
	where $\bar{\omega}^b$ and $\bar{v}^r$ are raw output of gyroscope and radar, respectively.
	Note that we consider the output of RANSAC/LSQ as the raw data from the radar (i.e., $\bar{v}^r = \hat{v}^r_{RL}$).
	The operator $\textrm{diag}\!\!: \mathbb{R}^n \rightarrow \mathbb{R}^{n\times n}$ returns a diagonal matrix.
	Next, let us assume that $n_g$ and $n_r$ are additive white Gaussian noises with zero mean and variance $\sigma^2_g$ and $\sigma^2_r$, respectively.
	Mathematical speaking, $n_g \thicksim {\mathcal{N}}\big(0_{3 \times 1}, \sigma^2_g I_3 \big)$ and $n_r \thicksim {\mathcal{N}}\big(0_{3 \times 1}, \sigma^2_r I_3 \big)$.
	Also, we assume that the gyroscope biases and radar scale factors are modeled as first-order Markov process model as \cite{dr}
	\begin{subequations}
		\begin{align}
			\dot{b}_g &= -\dfrac{1}{\tau_b} b_g + n_{b_g}, \quad
			\dot{s}_r = -\dfrac{1}{\tau_r} s_r + n_{s_r},
		\end{align}
	\end{subequations}
	where $\tau_b$ and $\tau_r$ are the time constants.
	Also, under the same assumption as stated above, we have $n_{b_g} \thicksim {\mathcal{N}}\big(0_{3 \times 1}, \sigma^2_{b_g} I_3 \big)$ and $n_{s_r} \thicksim {\mathcal{N}}\big(0_{3 \times 1}, \sigma^2_{s_r} I_3 \big)$.
	
	The time propagation of the quaternion vector $q_b^n$ is done by solving the following differential equation
	\begin{align} \label{quaternion_dynamic}
		\dot{q}_b^n = \dfrac{1}{2} q_b^n \otimes \big( \bar{\omega}^b - b_g \big).
	\end{align}
	Next, from \eqref{radar_mechanization}, the position differential equation is given by 
	\begin{align} \label{position_dynamic}
		\dot{p}_b^n = v_b^n.
	\end{align}
	Again, since error-state formulation is considered, we need to find the error dynamical equation.
	From \eqref{quaternion_error_def}, \eqref{gyroscop_measurement}, and with the error quaternion definition \eqref{quaternion_dynamic}.
	The procedure for determining the error dynamic model $\delta \dot{q}_b^n$ are well-established and can be found in \cite{sola2017quaternion} as
	\begin{align}
		\delta \dot{\Psi}_b^n &=  \hat{C}_b^n \delta b_g - \hat{C}_b^n n_g.
	\end{align}
	
	Now, from \eqref{radar_mechanization}, \eqref{error_def}, \eqref{cbn_error_def}, \eqref{radar_measurement}, and \eqref{position_dynamic}, one could derive the following differential equation
	\begin{align}
		\delta \dot{p}_b^n 
		&= \hat{C}_b^n C_r^b \textup{diag}( \bar{v}^r ) \delta s_r 
		+ \hat{C}_b^n C_r^b \textup{diag}(\hat{s}_r) n_r 
		+ \hat{C}_b^n \ssm{p_r^b} n_g \nonumber \\
		&\mathrel{\phantom{=}}- \ssm{\hat{C}_b^n \ssm{\hat{\omega}^b} p_{r}^b 
			- \hat{C}_b^n C_r^b \hat{v}^r} \delta \Psi_b^n 
		- \hat{C}_b^n \ssm{p_r^b} \delta b_g.
	\end{align}
	Here, we remove the hat symbol above the IMU-Radar calibration parameters $C_r^b$ and $p_r^b$ based on the assumption established in Section \ref{subsec:radar_mechanization} that makes them constant.
	
	Utilizing the derivation thus far, now we can formulate the linearized error dynamic equation of $\delta x_{dr}^n$ as follows:
	\begin{align}
		\delta x_{dr,k} = \Phi_{k-1} \delta x_{dr,k-1} + G_{k-1} w_{k-1}.
	\end{align}
	In this context, the process noise vector $w_{k-1}$ and its corresponding matrix  $G_{k-1}$ (for brevity, only non zero elements of $G_{k-1}$ are provided\footnote{\label{note1}Using C++ 0-based indexing notations.}) are respectively defined as  
	\begin{subequations}
		\begin{align}
			w_{k-1} &= \bbm n_r^\top & n_g^\top & n_{b_g}^\top & n_{s_r}^\top \ebm^\top \\
			G_{k-1}(0:2, 0:2) &= \hat{C}_b^n C_r^b \textup{diag}(\hat{s}_r) \\
			G_{k-1}(0:2, 3:5) &= \hat{C}_b^n \ssm{p_r^b} \\
			G_{k-1}(3:5, 3:5) &= -\hat{C}_b^n \\
			G_{k-1}(6:11, 6:11) &= I_6
		\end{align}
	\end{subequations}
	The transition matrix $\Phi_{k-1}$ is discretized as
	\begin{align}
		&\Phi_{k-1} = \textrm{exp} \big( F T \big) \nonumber \\
		&\approx I_{12} + F\big((k-1) T \big) T + 0.5 F\big((k-1) T \big)^2 T^2,
	\end{align}
	where $T$ denotes the radar sampling period, and the continuous-time error dynamics matrix $F \in \mathbb{R}^{12\times12}$ is given (for brevity, only non-zero elements of $F$ are provided\footref{note1})
	\begin{subequations}
		\begin{align}
			F(0:2, 3:5) &= - \ssm{\hat{C}_b^n \ssm{\hat{\omega}^b} p_{r}^b - \hat{C}_b^n C_r^b \hat{v}^r} \\
			F(0:2, 6:8) &= - \hat{C}_b^n \ssm{p_r^b}\\
			F(0:2, 9:11) &= \hat{C}_b^n C_r^b \textup{diag}( \bar{v}^r )\\
			F(3:5, 6:8) &=  \hat{C}_b^n \\
			F(6:8, 6:8) &= -\dfrac{1}{\tau_g} I_3, \,\, F(9:11, 9:11) = -\dfrac{1}{\tau_r} I_3.
		\end{align}
	\end{subequations}
	The augmented transition matrix is then given by
	\begin{align}
		\breve{\Phi}_{k-1} = \bbm \Phi_{k-1} & 0_{6 \times 6} \\ 0_{6 \times 6} & I_6 \ebm.
	\end{align}
	As a result, the error covariance of the augmented system is propagated as \cite{stochastic}
	\begin{align}
		\breve{P}_k^- = \bbm \Xi & \Upsilon(0:5, 0:5)^\top P_{k-1}^+ \\ 
		P_{k-1}^+ \Upsilon(0:5, 0:5) & P_{k-1}^+ \ebm,
	\end{align}
	where $\Xi = \Phi_{k-1} P_{k-1}^+ \Phi_{k-1}^\top + G_{k-1} Q_{d,k-1} G_{k-1}^\top$ with the discretized process noise covariance $Q_{d, k-1} = T \textrm{diag} \big(\sigma_g^2, \sigma_r^2, \sigma_{b_g}^2, \sigma_{s_r}^2\big)$ and $P_{k-1}^+$ is the \emph{a posteriori} error covariance of the DR error-state $\delta x_{dr, k-1}$.
	Note that $\Upsilon = \Pi_{i=1}^M \Phi_{k-i}$ accumulates over time while awaiting the measurement update step.
	
	\subsection{Measurement Model}
	The DeRO system leverages measurement from both radar (distance of the targets) and tilt angles calculated from accelerometers.
	We shall describe each measurement model in detail.
	\subsubsection{Radar range measurements}
	Given the relative distance $\bar{p}_{r,k}^{r,{k-M}}$ between the step $k$ and $k-M$ of the radar frame obtained from a scan matching algorithm.
	By solving a straightforward vector geometry problem, one could establish the following relationship
	\begin{align}
		\hat{p}_{r,k}^{r,{k-M}} &= C_b^{r} \hat{C}_n^{b,{k-M}} \big( \hat{p}_{b,{k}}^n - \hat{p}_{b,{k-M}}^n \big).
	\end{align}
	With this equation, the augmented residual $\breve{z}_{r,k}$ and its corresponding linearized measurement model is given by
	\begin{align}
		\breve{z}_{r, k} &= \hat{p}_{r,k}^{r,{k-M}} - \bar{p}_{r,k}^{r,{k-M}} = \breve{H}_{r,k} \delta x_k + n_{p_r},
	\end{align}
	where
	\begin{align}
		\breve{H}_{r,k} &= C_b^{r} \hat{C}_n^{b,{k-M}} \bbm I_3 & \ssm{\hat{C}_{b,k}^n p_r^b} & 0_{3\times6} & -I_3 & \Lambda \ebm,
	\end{align}
	$\Lambda \!=\! \ssm{\hat{p}_{b,{k-M}}^n - \hat{p}_{b,{k}}^n - \hat{C}_{b,k}^n p_r^b}$, and $n_{p_r} \!\!\thicksim {\mathcal{N}}\big(0_{3 \times 1}, \sigma^2_{p_r} I_3 \big)$.

	The process of deriving the aforementioned matrix can be referenced from \cite{stochastic}, with extensions provided for 3D cases.
	Unlike in \cite{radar-msckf}, where each target's distance measurement is directly used, we employ the entire scan for matching to determine the distance between frames.
	This strategy leads to a significant reduction in computational cost.
	\subsubsection{Tilt angles}
	It is well known that roll and pitch angles of the body frame can be computed from the accelerometer readings $\bar{f}^b$ without acceleration as \cite{groves}
	\begin{align}
		\bar{\phi}_a = \arctan\bigg( \dfrac{-\bar{f}_y}{-\bar{f}_z} \bigg), \,\, \bar{\theta}_a = \arctan\Bigg( \dfrac{\bar{f}_x}{\sqrt{\bar{f}_y^2 + \bar{f}_z^2}} \Bigg),
	\end{align}
	where $\bar{f}^b = \bbm \bar{f}_x & \bar{f}_y & \bar{f}_z \ebm^\top$.

	According to the relationship between Euler angles and misalignment angles \cite{jhj}, we establish the following linearized measurement model to update the DR estimation
	\begin{align}
		z_{a,k} &= \bbm \hat{\phi}_k \\ \hat{\theta}_k \ebm - \bbm \bar{\phi}_{a,k} \\ \bar{\theta}_{a,k}\ebm = H_{a,k} \delta \Psi_{b, k}^n \nonumber + n_a \\
		&= \bbm -\dfrac{\cos(\hat{\psi}_k)}{\cos(\hat{\theta}_k)} & -\dfrac{\cos(\hat{\psi}_k)}{\cos(\hat{\theta}_k)} & 0 \\[0.8em]
		\sin(\hat{\psi}_k) & -\cos(\hat{\psi}_k) & 0 \ebm \delta \Psi_{b, k}^n + n_a.
	\end{align}
	Here, the estimate Euler angle $\hat{\Psi}_{b,k}^n = \bbm \hat{\phi}_k & \hat{\theta}_k & \hat{\psi}_k \ebm^\top$ can be found by transforming the estimate quaternion $\hat{q}_{b,k}^n$ and $n_a \thicksim {\mathcal{N}}\big(0_{2 \times 1}, \sigma^2_{a} I_2 \big)$.
	Furthermore, given that velocity information is available from radar measurements, we compensate for the linear acceleration and biases derived from accelerometer output.
	This compensation enhances the quality of tilt angle information.
	Particularly,
	\begin{align} \label{accel_compen}
		\hat{f}^b_k = \bar{f}^b_k - \hat{b}_{a,k} - \hat{C}_n^{b,k} \dfrac{\big( \hat{v}_{b,k}^n - \hat{v}_{b,{k-1}}^n \big)}{T}.
	\end{align}
	It is important to note that accelerometer sensors are assumed to be pre-calibrated by a separate algorithm (e.g., coarse alignment) to obtain $\hat{b}_a$.
	To further prevent adverse effects of linear acceleration and biases, we adopt the adaptive rule proposed by Park \textit{et al.} \cite{dr}.
	More specifically, we compare the difference $\Vert \hat{f}^b \Vert - \Vert g \Vert$, where $g$ represents the gravitational force vector, with a predetermined constant threshold $\gamma$.
	If this difference exceeds $\gamma$, we increment the value of $\sigma^2_a$.
	This approach has been effectively employed in \cite{jhj}.
	
	\subsection{Implementation}
	The detailed step-by-step practical implementation of the DeRO is outlined in Algorithm \ref{algo}.
	Here, we assume accelerometer data is collected faster than radar data, allowing approximate timestamp synchronization.
	This enables both data sources to be used concurrently for KF's measurement update.
	Furthermore, to enhance the performance of ICP algorithm, we utilize the estimations from the previous $M$ steps to calculate the initial guessed pose.

	\begin{algorithm}
		\caption{DeRO} \label{algo}
		\begin{algorithmic}[1]
			\State \textbf{Initialize:} $\hat{x}_{dr, 0}^+, P^+_0, M, \sigma_g^2, \sigma_{b_g}^2, \sigma_{s_r}^2, \sigma_a^2, scanCount$.
			\State Performing coarse alignment algorithm $\rightarrow \hat{b}_g, \hat{b}_a$.
			\For {$k = 1$ to $K$}
			\If {Gyroscope is available}
			\State Compensating gyroscope biases $\hat{\omega}^b = \bar{\omega}^b - \hat{b}_g$.
			\State Gyroscope mechanization.
			\EndIf
			\If {Radar is available}
			\State Increase $scanCount$ by 1. 
			\State RANSAC/LSQ $\rightarrow \hat{v}^r, \hat{\sigma}_r^2$.
			\State Accumulating $\Upsilon$.
			\State Radar-based dead reckoning mechanization.
			\State Time update $\rightarrow \breve{\Phi}_{k-1}, \breve{P}_k^-$.
			\If {$scanCount = M$}
			\State Registration $\rightarrow \hat{p}_{r,k}^{r,{k-M}}, \hat{\sigma}_{p_r}^2$.
			\State Computing $\breve{H}_{r}$.
			\If {Accelerometer is available}
			\State Compensating accelerometer bias and 
			\Statex  \quad \quad \quad \quad \quad \quad linear acceleration $\leftarrow$ \eqref{accel_compen}.
			\State Calculating $\rightarrow \hat{\phi}_{a}, \hat{\theta}_{a}$ and $H_a$.
			\State Stacking $H_a$ and $\breve{H}_{r}$.
			\EndIf
			\State Estimation update with stacked measurement 
			\Statex \quad \quad \quad \quad \, matrix and $\breve{P}_k^-$.
			\State Reset $scanCount$.
			\EndIf
			\EndIf
			\EndFor
		\end{algorithmic}
	\end{algorithm}
	
	\section{EXPERIMENTS} \label{sec:results}
	\subsection{Open-source datasets}
	To verify the proposed DeRO algorithm, five open source datasets provided by \cite{doer-yaw} was used, namely Carried 1 to 5.
	We briefly describe overview of the datasets; for more detailed information about the setup, readers are encouraged to refer to the mentioned paper.
	These datasets were recorded in an office building using a hand-held sensor platform equipped with a 4D milimeter-wave FMCW radar (TI IWR6843AOP) and an IMU (Analog Devices ADIS16448).
	The IMU provides data at approximately 400 Hz, while the radar captures 4D point clouds at 10 Hz.
	In addition, the radar has a field of view of 120 deg for both azimuth and elevation angles. 
	Furthermore, the sensor platform includes a monocular camera, which is utilized to generate a pseudo ground truth via the visual inertial navigation system (VINS) algorithm with loop closure.
	%
	
	We implemented DeRO in ROS 2 Foxy \cite{ros2} using C++.
	The datasets were processed on a Linux PC equipped with an Intel i9-12900K CPU running at 3.20 GHz.
	Additionally, to ensure fairness in comparison, we independently replicated the EKF RIO instead of relying on the provided code by the authors.
	Note that we maintained consistency with the algorithm's parameters, such as RANSAC/LSQ, sensor noise density, initialization values, etc., as specified in \cite{DoerMFI2020}.
	For scan matching, we employed a standard iterative closest point (ICP) algorithm with the window size $M =$ 3.
	The rationale behind this number is that the radar measurement frequency in the public dataset is approximately 10 Hz.
	We aim for the filter to update the estimation twice per second.
	Based on the quality metric provided by the ICP output, we manually tuned the measurement noise variance $\sigma_{p_r}^2$.
	Besides, we selected $\gamma =$ 0.059 m/s$^2$, as recommended by \cite{dr}, for the noise covariance adaptive rule.

	\subsection{Evaluation}
	For evaluation purposes, we aligned the estimated trajectories with the pseudo ground truth using the method described in \cite{eval}, employing position-yaw alignment.
	This tool also calculates the absolute trajectory error (ATE) metric, as defined in \cite{kitti}, which we will utilize for comparison.
	\begin{figure}[thpb]
		\centering
		\includegraphics[scale=0.585]{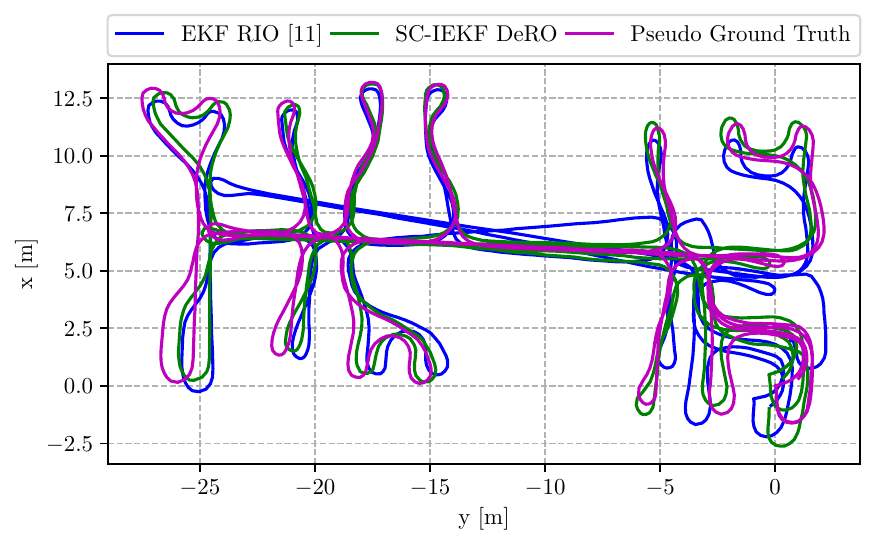}
		\caption{Comparison of 2D trajectory estimation between EKF RIO and SC-IEKF DeRO of the Carried 1 dataset.}
		\label{2D_trajectory}
	\end{figure}

	\subsubsection{Evaluation of 2D Trajectory}
	Figure \ref{2D_trajectory} illustrates the top-view of the aligned X-Y trajectory estimates obtained by the EKF RIO and SC-IEKF DeRO.
	Overall, Both methods demonstrate decent estimation performance.
	However, our proposed method exhibits superior performance, particularly evident when the sensor platform traces a straight loop from $(x,y)=(7,0)$ to $(x,y) = (7,-25)$.
	Furthermore, as the start and end points of the trajectory coincide, we can relatively judge the filter's accuracy solely by examining the final estimated positions.
	Specifically, SC-IEKF DeRO yields an approximate position of $(\hat{x}, \hat{y}, \hat{z}) = (-1.07, -0.29, 0.75)$, while EKF RIO yields approximately $(3.15, -1.37, 1.34)$, corresponding to distance errors of approximately 1.3 m and 3.7 m, respectively.
	%
	%
	Thus, our proposed method reduces the final position distance error from the conventional approach by approximately 65\%.
	
	\begin{figure}[b]
		\centering
		\includegraphics[scale=0.585]{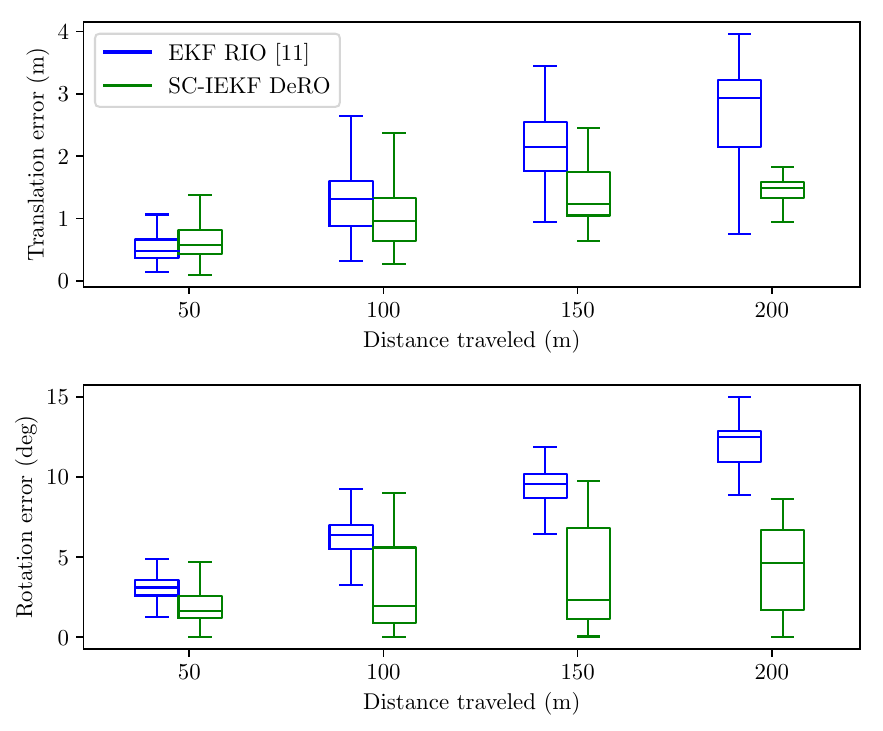}
		\caption{Boxplots of the relative translation and rotation errors over distance traveled between the two investigated method of the Carried 1 dataset.}
		\label{trans_rot}
	\end{figure}

	\subsubsection{Evaluation of Relative Errors}
	The relative errors in terms of translation and rotation are depicted in Fig. \ref{trans_rot} using boxplots.
	Our method consistently outperforms the EKF RIO in both categories.
	Notably, although the translation error of SC-IEKF DeRO shows slightly higher error and uncertainty after the sensor platform travels 50 m, it reliably maintains the error within 2.5 m thereafter.
	In contrast, the EKF RIO experiences a rapid position drift, reaching up to 4 m over a distances of 200 m.
	For rotation estimation, while the proposed algorithm demonstrates a smaller mean, it exhibits a larger interquartile range due to the slower update rate of the ICP compared to the traditional method.

	\begin{figure}[thpb]
		\centering
		\includegraphics[scale=0.19]{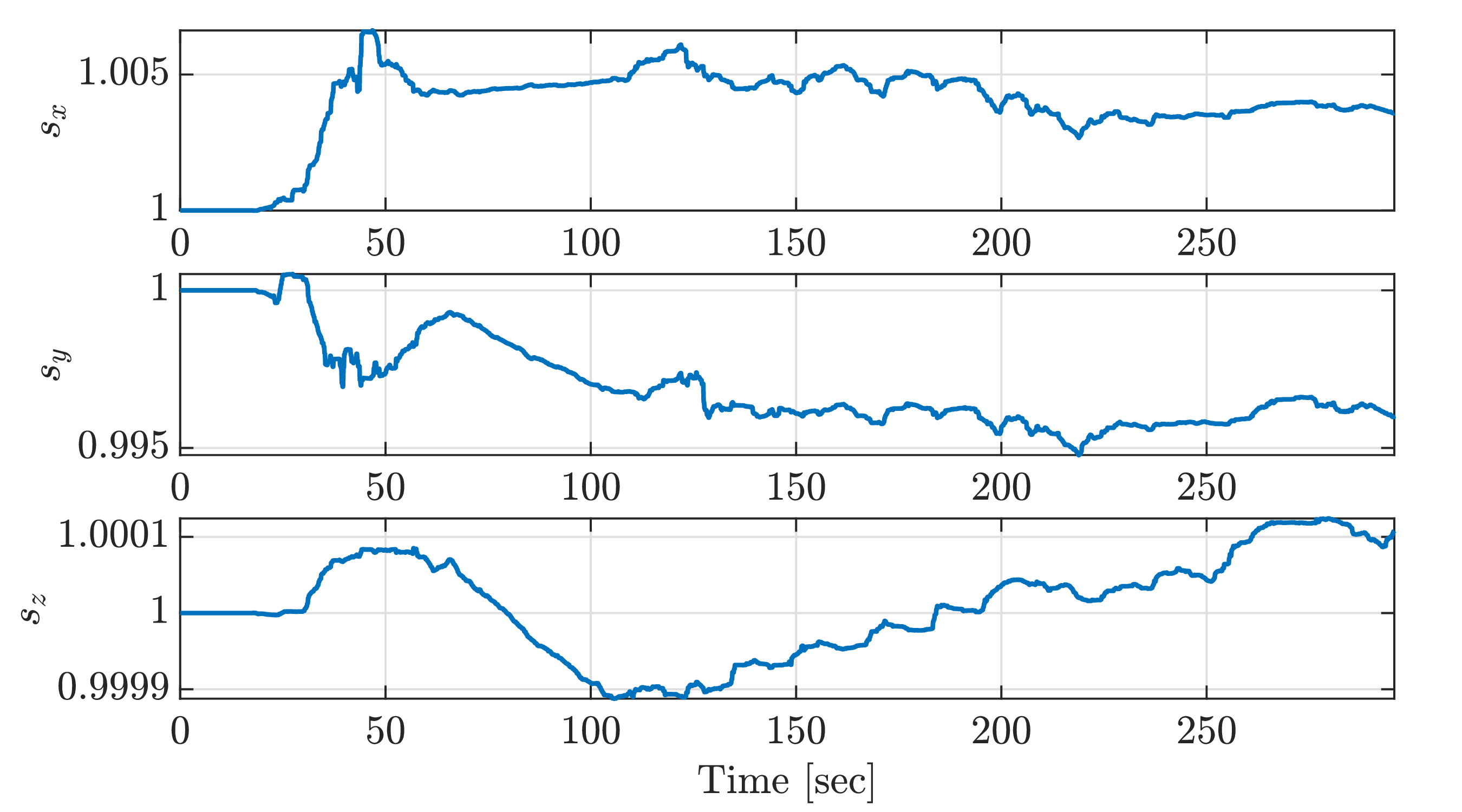}
		\caption{3D scale factor estimation of the proposed method with the Carried 1 dataset.}
		\label{scale}
	\end{figure}

	Fig. \ref{scale} reports the radar scale factor estimation obtained using the presented approach.
	From the plot, it is evident that the scale factor along the $x$-axis tends to converge and stabilize at 1.005, while it stabilizes at 0.995 along the $y$-axis, and approximately 1 with small fluctuations along the $z$-axis.
	This observation, coupled with the analysis of the estimation results, supports our belief that compensating for radar scale factor is crucial and can lead to significant improvements.

	\renewcommand{\arraystretch}{1.2}
	\begin{table}[h]
		\caption{Absolute trajectory error in terms of translation and rotation across five open-source datasets, with trajectory length and total time provided, for the two considered methods.}
		\label{table}
		\begin{center}
			\begin{tabular}{ccc cc}
				\toprule[1pt]
				\multirow{2}{*}{Dataset} & \multicolumn{2}{c}{EKF RIO \cite{DoerMFI2020}} & \multicolumn{2}{c}{SC-IEKF DeRO}\\
				\cmidrule(lr){2-3}\cmidrule(lr){4-5}
				& [m] & [deg] & [m] & [deg] \\
				
				\cmidrule(lr){1-1} \cmidrule(lr){2-3} \cmidrule(lr){4-5}
				Carried 1 (287 m, 273 s) & $1.075$ & $5.370$  & $\boldsymbol{0.617}$ & $\boldsymbol{2.734}$ \\
				Carried 2 (451 m, 392 s) & $2.724$ & $14.768$ & $\boldsymbol{0.709}$ & $\boldsymbol{2.869}$ \\
				Carried 3 (235 m, 171 s) & $1.110$ & $5.741$  & $\boldsymbol{0.464}$ & $\boldsymbol{1.853}$ \\
				Carried 4 (311 m, 220 s) & $1.547$ & $7.448$  & $\boldsymbol{0.491}$ & $\boldsymbol{2.659}$ \\
				Carried 5 (228 m, 172 s) & $1.495$ & $5.331$  & $\boldsymbol{0.741}$ & $\boldsymbol{3.093}$ \\
				\cmidrule(lr){1-1} \cmidrule(lr){2-3} \cmidrule(lr){4-5}
				Mean ATE & $1.590$ & $7.731$ & $\boldsymbol{0.604}$ & $\boldsymbol{2.641}$ \\
				\bottomrule[1pt]
				\multicolumn{5}{l}{\footnotesize The bold numbers represent better results (the smaller number).}
			\end{tabular}
		\end{center}
	\end{table}

	\subsubsection{Evaluation of ATE}
	Table \ref{table} summarizes the ATE in terms of both translation and rotation for the two considered algorithms across all datasets.
	Once again, the SC-IEKF DeRO completely outperforms the EKF RIO in all trials, especially in the Carried 1 and 2 datasets, which features numerous turns.
	For instance, in the Carried 2 field test (the longest trajectory with 451 m distance and 392 seconds duration), our developed method yields a translation error of 0.709 m and rotation error of 2.869 deg, compared to 2.724 m and 14.768 deg for the EKF RIO.
	This represents a translation error reduction of 74\% and a rotation error reduction of 80\%.
	Overall, across the mean ATE of the five experiments, our DeRO approach reduces the translation error by approximately 62\% and the rotation error by 66\%.
	\section{CONCLUSION} \label{sec:conclusion}
	In this article, we have proposed DeRO, a framework of dead reckoning based on radar odometry with tilt angle measurements from accelerometer.
	In contrast to the previous studies where radar measurements are used solely to update the estimations, we employ the 4D FMCW radar as a hybrid component of our system.
	Specifically, we leverage Doppler velocity obtained by the RANSAC/LSQ algorithm with gyroscope measurements to primarily perform dead reckoning (calculating poses).
	The radar's range measurements, in conjunction with accelerometer data, are utilized for the filter's measurement update.
	This approach enables estimation and compensation of Doppler velocity scale factor errors, while also allowing for compensation of linear acceleration from accelerometers using radar velocity estimation.
	Moreover, we apply the renowned stochastic cloning-based IEKF to address relative distance problem obtained from the scan matching.
	The effectiveness of our proposed method is validated through a comprehensive evaluation using a set of open-source datasets.
	As expected, directly utilizing radar velocity instead of integrating acceleration offers a significantly improved odometry solution.
	The provided mean ATE across all test fields demonstrates that SC-IEKF DeRO achieves substantially better overall performance compared to its competitor.
	One limitation of this study is that the DeRO approach operates at a relatively slow rate due to the radar component.
	Addressing this limitation will be a focus of our future research.
	
	


	
	

	
	\bibliographystyle{IEEEtran}
	\bibliography{IEEEabrv,iros_viet_bib}
	
\end{document}